\PassOptionsToPackage{table,dvipsnames}{xcolor}
\documentclass[10pt,twocolumn,letterpaper]{article}

\usepackage[pagenumbers]{cvpr} % To force page numbers, e.g. for an arXiv version

\usepackage{multirow}
\usepackage{arydshln}
\usepackage{algorithm}
\usepackage{algorithmic}
\usepackage{pifont}
\usepackage[accsupp]{axessibility}

\definecolor{cvprblue}{rgb}{0.21,0.49,0.74}
\usepackage[pagebackref,breaklinks,colorlinks,allcolors=cvprblue]{hyperref}

\def\paperID{990} % *** Enter the Paper ID here
\def\confName{CVPR}
\def\confYear{2025}

\newcounter{eqfncorr}
\def\equalcorr{
  \ifnum\value{eqfncorr}=0
    \thanks{Corresponding Authors}
    \setcounter{eqfncorr}{\value{footnote}}
  \else
    \footnotemark[\value{eqfncorr}]
  \fi
}

\title{Boosting Domain Incremental Learning: \\Selecting the Optimal Parameters is All You Need}

\author{Qiang Wang\textsuperscript{\rm 1}, 
Xiang Song\textsuperscript{\rm 1}\equalcorr, 
Yuhang He\textsuperscript{\rm 1}\equalcorr, 
Jizhou Han\textsuperscript{\rm 1}, 
Chenhao Ding\textsuperscript{\rm 1},
Xinyuan Gao\textsuperscript{\rm 1},
Yihong Gong\textsuperscript{\rm 1,2}\\
\textsuperscript{\rm 1}Xi’an Jiaotong University
\textsuperscript{\rm 2}Shenzhen University of Advanced Technology\\
{\tt\small \{qwang,songxiang\}@stu.xjtu.edu.cn, heyuhang@xjtu.edu.cn} \\
{\tt\small \{jizhou\_han,dch225739,gxy010317\}@stu.xjtu.edu.cn, ygong@mail.xjtu.edu.cn}
}

\begin{document}
\maketitle

\begin{abstract}
Deep neural networks (DNNs) often underperform in real-world, dynamic settings where data distributions change over time. Domain Incremental Learning (DIL) offers a solution by enabling continual model adaptation, with Parameter-Isolation DIL (PIDIL) emerging as a promising paradigm to reduce knowledge conflicts. However, existing PIDIL methods struggle with parameter selection accuracy, especially as the number of domains and corresponding classes grows. To address this, we propose SOYO, a lightweight framework that improves domain selection in PIDIL. SOYO introduces a Gaussian Mixture Compressor (GMC) and Domain Feature Resampler (DFR) to store and balance prior domain data efficiently, while a Multi-level Domain Feature Fusion Network (MDFN) enhances domain feature extraction. Our framework supports multiple Parameter-Efficient Fine-Tuning (PEFT) methods and is validated across tasks such as image classification, object detection, and speech enhancement. Experimental results on six benchmarks demonstrate SOYO's consistent superiority over existing baselines, showcasing its robustness and adaptability in complex, evolving environments. The codes will be released in \href{https://github.com/qwangcv/SOYO}{https://github.com/qwangcv/SOYO}.
\end{abstract}
\vspace{-0.2cm}

\section{Introduction}

\begin{figure}[t]
  \centering
  \includegraphics[width=0.42\textwidth]{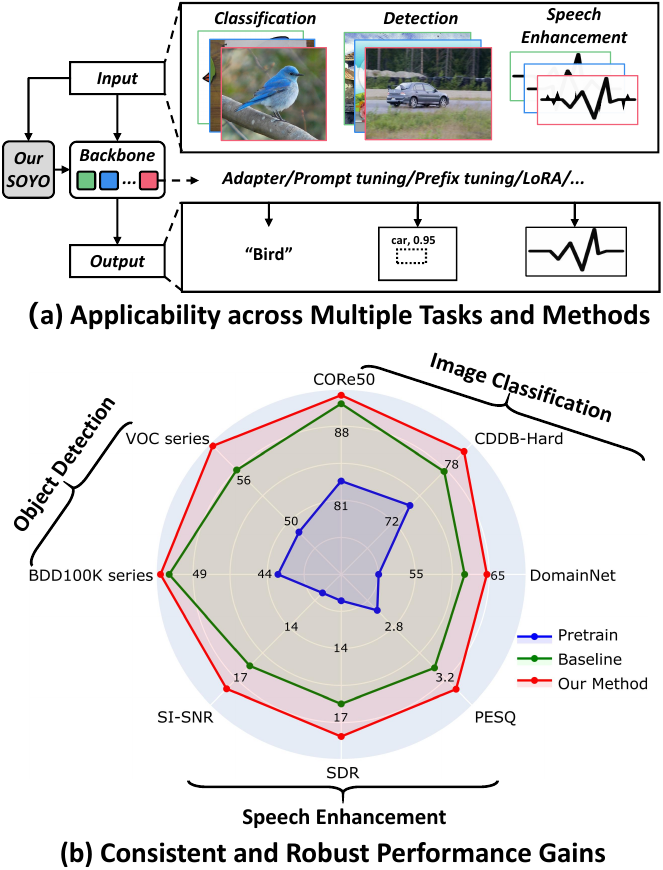}
  \caption{\textbf{Illustration of the proposed SOYO.} (a) The green, blue, and red small squares represent learnable parameters for different domains. (b) The ``pretrain'' represents evaluation using the pre-trained model directly. The lower bound of the radar chart is set to 90\% of the ``pretrain'', while the upper bound represents the achievable upper limit of the baseline if the domain labels of test samples are known in advance. Best viewed in color.}
  \label{fig:intro_fig}
  \vspace{-0.4cm}
\end{figure}

Deep neural networks (DNNs) have achieved significant success in various fields such as image recognition, object detection, and speech recognition. However, existing methods are typically designed for static datasets or controlled environments, making them less adaptable to dynamic real-world settings. In practical applications, target objects of interest may be affected by occlusion, appearance changes, and fluctuations in environmental conditions, leading to continuously evolving data distributions. For instance, autonomous vehicles need to operate safely under various conditions, including daytime, nighttime, rain, and fog. Current models struggle to handle such substantial domain changes effectively. To address this challenge, Domain Incremental Learning (DIL) has received widespread attention~\cite{wang2023isolation,lamers2023clustering,wang2024multi,shi2024unified,chen2024leaving,ding2025space} in recent years, which requires models to continuously adapt to new data distributions while retaining knowledge of previously learned distributions.

Classical DIL methods generally employ knowledge distillation~\cite{hinton2015distilling,li2017learning,rebuffi2017icarl,hou2019learning,li2022learning,kang2022class,ding2023domain,song2024overcoming,han2025learn} or parameter regularization~\cite{kirkpatrick2017overcoming,zenke2017continual,aljundi2018memory,akyurek2021subspace,liu2022few,shi2023multi,dong2023knowledge} techniques to constrain model parameter updates, thereby preventing overfitting to new domains and alleviating catastrophic forgetting. However, these methods inevitably involve a trade-off between retaining old knowledge and learning new knowledge, where improvements in one domain often lead to reduced accuracy in another, ultimately limiting overall performance. To address this challenge, a series of DIL methods based Parameter-Isolation (PIDIL) have been proposed, which allocate and train distinct parameters to fine-tune pre-trained models for different domains, aiming to optimize performance for each domain individually. During inference, these methods predict the domain labels of the test samples and select the corresponding parameters for these samples. PIDIL has demonstrated its effectiveness in preventing conflicts between new and old knowledge, achieving optimal performance in image classification~\cite{sprompt,cprompt,PINA}, object detection~\cite{LDB}, and speech enhancement~\cite{LNA} tasks.

The strong performance of PIDIL is attributed to two main factors: the fine-tuning strategy for learning new domains and the accuracy of parameter selection during the inference stage. Existing methods~\cite{garg2022multi,l2p,dualprompt,coda,cprompt,PINA,gao2024beyond,dualcp} primarily focus on designing fine-tuning strategies to effectively capture domain-specific information, such as prompt tuning, prefix tuning, and adapters. A few approaches~\cite{nicolas2024mop,LDB,PINA} have recognized the impact of parameter selection accuracy on PIDIL performance, employing various domain prediction methods like k-nearest neighbors, nearest mean classifier, and patch shuffle selector. However, as the number of domains increases, the domain prediction accuracy of these strategies inevitably declines, resulting in inaccurate domain-specific information being applied during inference and limiting overall performance. Moreover, these studies only evaluated single-task scenarios, leaving their effectiveness across multiple tasks unverified. Therefore, it is crucial to design a general and robust domain prediction method to improve parameter selection accuracy, thereby enhancing the performance of state-of-the-art PIDIL methods across various downstream tasks.

In the paper, we propose a trainable and lightweight framework named SOYO to select the optimal parameters for PIDIL. The SOYO focuses on identifying the domain label and is trained using the domain features of the samples. However, the data from previous domains is inaccessible in DIL, which leads to class imbalance issues in SOYO training. To address this problem while minimizing memory usage and protecting data privacy, we design a Gaussian Mixture Compressor (GMC) that models the previous domain features as a linear combination of several Gaussian distributions, storing only the parameters of these distributions. Then, we use a Domain Feature Resampler (DFR) to reconstruct pseudo-domain features to simulate the original feature distribution. By random sampling from the pseudo-domain features and the current domain features, we effectively mitigate the class imbalance problem in SOYO training. In addition, we design a Multi-level Domain Feature Fusion Network (MDFN) to extract domain features more effectively. Specifically, the shallow layers of the backbone capture lower-level spatial information, while the deep layers extract global semantic features. Existing works~\cite{sprompt,LDB,PINA} typically use only the features output by the final layer as the domain features, which limits the diversity of the learned representations. In contrast, our MDFN fuses the spatial and semantic features to obtain more discriminative domain features.

Furthermore, our framework is robust and applicable to all PIDIL methods, including image classification, object detection, and speech enhancement (see \cref{fig:intro_fig}~(a)). It also supports various Parameter-Efficient Fine-Tuning (PEFT) methods, such as Adapter, prompt tuning, prefix tuning, \emph{etc}. We evaluate our method on six benchmarks and across eight metrics as illustrated in \cref{fig:intro_fig}~(b). Experimental results show that our method effectively boosts existing baselines, achieving state-of-the-art performance. Our contributions can be summarized as follows:

\begin{itemize}
\item We propose SOYO, a novel and lightweight framework for domain label prediction. The framework enhances parameter selection accuracy during inference, boosting model performance across various tasks in Parameter-Isolation Domain Incremental Learning (PIDIL).

\item We introduce a Gaussian Mixture Compressor (GMC) and a Domain Feature Resampler (DFR) to balance the training of the SOYO, where the GMC efficiently stores the key information of previous domains and the DFR simulates the original feature distribution without increasing memory usage or compromising data privacy.

\item We develop a Multi-level Domain Feature Fusion Network (MDFN) to extract discriminative domain features by fusing spatial information with semantic information, improving the accuracy of domain label prediction.

\item Our SOYO framework is compatible with a wide range of DIL tasks and parameter-efficient fine-tuning methods. Extensive evaluations on six benchmarks demonstrate that our approach consistently outperforms existing PIDIL baselines and achieves SOTA performance.
\end{itemize}

\section{Related Work}
\subsection{Domain Incremental Learning}
Many visual and audio tasks encounter challenges of catastrophic forgetting due to variations in input data. To validate the robustness of our approach, we conducted experiments on three DIL tasks: Domain Incremental Classification (DIC), Domain Incremental Object Detection (DIOD), and Domain Incremental Speech Enhancement (DISE).

\textbf{DIC.} DIC is the most fundamental task in DIL and has received the most exploration. S-Prompts~\cite{sprompt} advocate for using prompt tuning to capture domain-specific information. Method~\cite{wang2024importance} proposes a shared parameter subspace learning approach, where parameters are updated using a momentum-based method. Methods~\cite{l2p,dualprompt,coda} combine learned parameters to handle test samples. The approach~\cite{wistuba2024choice} explores the effectiveness of LoRA strategies.

\textbf{DIOD.} Object detection is a critical task in computer vision. The methods~\cite{liu2020multi,liu2023continual,yang2023one} use exemplar storage and knowledge distillation strategies to mitigate catastrophic forgetting. CIFRCN~\cite{cifrcn} achieves non-exemplar DIOD by extending the region proposal network, while ERD~\cite{erd} proposes a response-based approach for transferring category knowledge. LDB~\cite{LDB} introduces a bias-tuning method to learn domain-specific biases, enabling incremental detection of objects across different domains.

\textbf{DISE.} In speech communication and automatic speech recognition systems, noise often interferes with speech, lowering audio perceptual quality and increasing the risk of misunderstanding. It is essential to study speech enhancement for reducing noise interference~\cite{xu2013experimental,yuliani2021speech,xu2024improving}. To adapt to various types of noise, LNA~\cite{LNA} proposes an adapter-based approach to address incremental speech enhancement and achieve strong performance.

\subsection{Parameter Selection Strategies on PIDIL}
In Parameter-Isolation Domain Incremental Learning (PIDIL), a set of additional parameters is trained to fine-tune the pre-trained model for each incremental domain. This paradigm ensures that information from all domains is preserved during training. However, domain labels are unknown at the inference stage. Therefore, PIDIL methods must design a domain label prediction strategy, \emph{i.e.}, parameter selection strategy for the test samples.

S-Prompts~\cite{sprompt} predicts domain labels by applying KMeans to cluster training features and using K-Nearest Neighbors (KNN) to match test samples with the nearest cluster center. LDB~\cite{LDB} simplifies this process by adopting a nearest mean classifier, equivalent to KNN with $k=1$. MoP-CLIP~\cite{nicolas2024mop} explores various distance metrics to improve feature clustering, including L1, L2, and Mahalanobis distances. PINA~\cite{PINA} uses a patch shuffle selector to disrupt class-dependent information and enhance domain feature extraction. These methods are training-free and easy to implement. However, their performance tends to decline as the number of domains increases. In contrast, the proposed SOYO is a training-based approach that selects parameters more accurately using only a few learnable parameters, leading to consistent performance improvements across diverse tasks.

\begin{figure*}[t]
  \centering
  \includegraphics[width=0.95\textwidth]{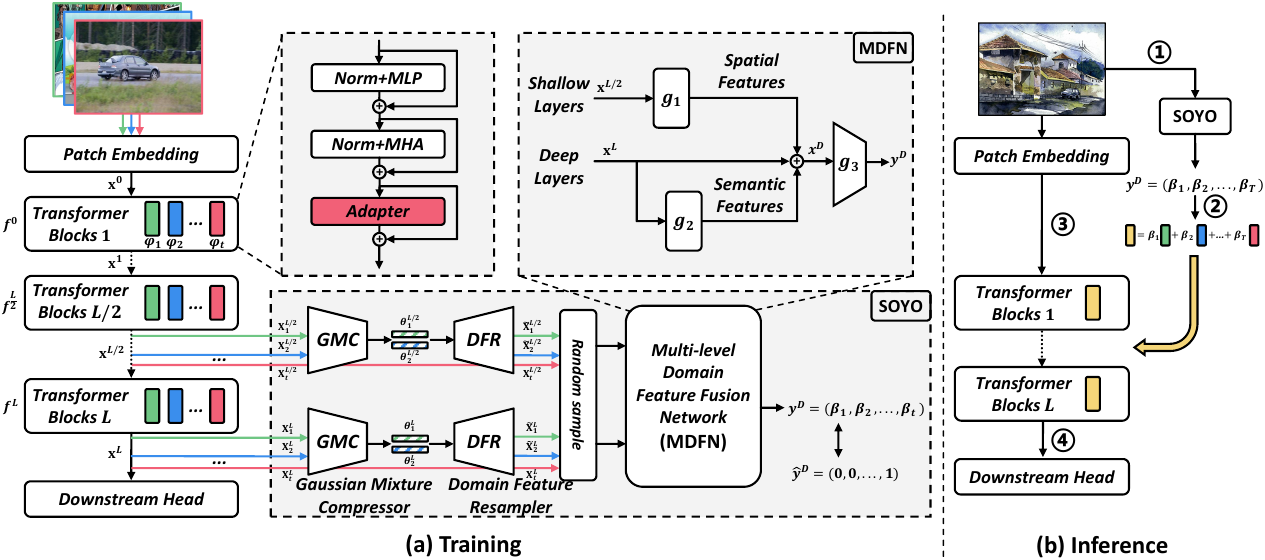}
  \caption{\textbf{Illustration of the proposed framework.} The numbers 1, 2, 3, and 4 in (b) indicate the sequence of steps. Best viewed in color.}
  \label{fig:framework}
\end{figure*}

\section{Method}
\subsection{Problem Formulation}
Given a deep learning task with an input space $\mathcal{X}$, output space $\mathcal{Y}$, and an ideal mapping function $f: \mathcal{X} \to \mathcal{Y}$, the nature of $\mathcal{Y}$ varies depending on the task. For an image classification task, $\mathcal{Y}$ represents a set of discrete labels. In an object detection task, $\mathcal{Y}$ corresponds to a set of object bounding boxes and their categories. In a speech enhancement task, $\mathcal{X}$ and $\mathcal{Y}$ represent raw audio segments and denoised audio segments, respectively.

Domain Incremental Learning (DIL) aims to continuously train and fine-tune a model to adapt to data from new domains while retaining knowledge from previous domains. Specifically, let $\{D_1, D_2, \dots, D_T\}$ represent a data sequence from the first domain to the $T$-th domain. The data $D_t$ from each domain contains a training set $(\mathcal{X}_t, \mathcal{Y}_t)$ and a test set $(\mathcal{X}'_t, \mathcal{Y}'_t)$. The objective is to design a model that can learn incrementally across domains, performing well on the new domain $D_t$ without significantly degrading performance on the previous domains $\{D_1, D_2, \dots, D_{t-1}\}$.

\subsection{Overall Framework}
\label{sec:framework}

\cref{fig:framework} illustrates the framework of our proposed method, which consists of two phases: (a) training and (b) inference. We use a visual task to demonstrate our approach. In the domain incremental setting, the backbone processes images from different domains, represented by green (first domain), blue (second domain), and red ($t$-th domain). Images are divided into patches and passed through a patch embedding layer to generate initial feature representations $\mathbf{x}^{0}$. These features are then processed by $L$ stacked transformer blocks, denoted as $f^{1}, f^{2}, \cdots, f^{L}$. The output feature of the $l$-th layer is denoted as $\mathbf{x}^{l}=f^{l}(\mathbf{x}^{l-1};\delta^{l},\phi^{l})$. Each block includes a Multi-Head Attention (MHA) mechanism, a Multilayer Perceptron (MLP), and a Layer Normalization (Norm) for stability, with the parameters $\delta^{l}$. Additional trainable parameters $\phi^{l}$ are introduced in each transformer block to capture domain-specific knowledge. The strategy to add the parameters $\phi^{l}$ can be implemented in various ways, such as prompt tuning~\cite{sprompt} or adapter~\cite{PINA}. Here we present the adapter and denote these domain-specific parameters as $\phi_{t}$ for the $t$-th domain.

Based on this backbone, we propose the SOYO method as a parameter selector, \emph{i.e.}, a domain label predictor. Unlike training-free KNN or NMC methods, SOYO is a trainable approach that effectively captures domain-discriminative features. It consists of a Gaussian Mixture Compressor (GMC), a Domain Feature Resampler (DFR), and a Multi-level Domain Feature Fusion Network (MDFN). The GMC compresses and retains critical information in the domain feature distribution, while the DFR reconstructs pseudo-domain features to enhance feature diversity. The MDFN is a lightweight trainable network that fuses features from shallow and deep layers to produce more discriminative domain features and obtain the domain prediction result. In the inference phase, the SOYO predicts the domain label of the input image and integrates the corresponding parameters into the transformer blocks as shown in \cref{fig:framework} (b). Please refer to Algorithm 1 in the appendix for more details.

\subsection{Learn to Select the Optimal Parameters}
\label{sec:MDFN}

Domain labels are unknown in the inference phase of DIL. To predict the domain label of a test sample and select the optimal parameters, we propose the Multi-level Domain Feature Fusion Network (MDFN) to extract domain features. Specifically, we extract features from the $\frac{L}{2}$-th and $L$-th layers of the model, denoted as $\mathbf{x}^{\frac{L}{2}}$ and $\mathbf{x}^{L}$, respectively. The shallow features capture low-level spatial features and color information, while the deep features focus on global semantic information. To balance shallow spatial information with deep semantic information, we use $\mathbf{x}^{\frac{L}{2}}$ as auxiliary information to help the MDFN predict the domain label as shown in the MDFN box of \cref{fig:framework} (a). The fused domain features $\mathbf{x}^{D} \in \mathbb{R}^{1 \times d}$ are computed as follows:
\begin{equation}
    \mathbf{x}^{D}=\mathbf{x}^{L}+g_1(\mathbf{x}^{\frac{L}{2}};\delta_{1})+g_2(\mathbf{x}^{L};\delta_{2}),
    \label{eq:MDFN}
\end{equation}
where $d$ represents the feature dimension, $g_1$ encodes information from the intermediate layer feature $\mathbf{x}_{\frac{L}{2}}$, and $g_2$ further refines the information from $\mathbf{x}_{L}$. Both $g_1$ and $g_2$ are Multilayer Perceptrons (MLP).

After obtaining the fused domain feature $\mathbf{x}_{D}$, we use a domain classification head $g_3$ to map it to a predicted domain probability vector $\mathbf{y}^{D}=g_{3}(\mathbf{x}^{D};\delta_{3})$, where $\mathbf{y}^{D} \in \mathbb{R}^{1 \times t}$ and $t$ is the total number of the known domains. For the $t$-th domain, the ground truth of the domain label is represented as a one-hot vector $\hat{\mathbf{y}}^{D}$, \emph{i.e.}, $\hat{\mathbf{y}}^{D}_{t}=1$ and all other elements are 0. The MDFN is trained using cross-entropy loss, formulated as follows:
\vspace{-0.2cm}
\begin{equation}\label{eq;loss}
    \mathcal{L}(\mathbf{y}^{D}, \hat{\mathbf{y}}^{D}) = - \sum_{i=1}^{t} \hat{\mathbf{y}}^{D}_i \log(\mathbf{y}^{D}_i).
    \vspace{-0.2cm}
\end{equation}
We optimize the parameters $\Delta = \{\delta_{1}, \delta_{2}, \delta_{3}\}$ by minimizing $\mathcal{L}(\mathbf{y}^{D}, \hat{\mathbf{y}}^{D})$.

\subsection{Balance MDFN Training}

\textbf{Gaussian Mixture Compressor.}
When the $t$-th domain arrives, we use data from domain $t$ to train the MDFN. However, training only with features from domain $t$ can lead to class imbalance, where the model tends to predict the domain label of test images as $t$. Nonetheless, storing all previous domain features is impractical due to storage limitations and privacy constraints. To address this issue, we introduce a Gaussian Mixture Compressor (GMC) to compress features from domains $1$ to $t-1$, retaining the most essential information to balance the MDFN training. Specifically, assume that the $\tau$-th domain contains $N_\tau$ samples, and the features extracted by the $l$-th transformer block be denoted as $\mathbf{X}^{l}_{\tau} = \{\mathbf{x}^{l}_{\tau,i}\}_{i=1}^{N_\tau} \in \mathbb{R}^{N_\tau\times d}$, where $d$ is the feature dimension. We model these features as a mixture of $K$ Gaussian distributions. The probability density function (PDF) of a single $d$-dimensional Gaussian distribution is given by:
\begin{equation}
\footnotesize
    \mathcal{N}(\mathbf{x}|\mu, \Sigma) = \frac{1}{(2\pi)^{\frac{d}{2}}|\Sigma|^{\frac{1}{2}}} \exp\left(-\frac{1}{2}(\mathbf{x} - \mu)^T \Sigma^{-1} (\mathbf{x} - \mu)\right),
\end{equation}
where $\mu$ is the mean vector, and $\Sigma$ is the covariance matrix. The PDF of the mixture of $K$ Gaussian distributions is represented as:
\vspace{-0.2cm}
\begin{equation}
    p(\mathbf{x} | \theta) = \sum_{k=1}^{K} \lambda_k \mathcal{N}(\mathbf{x} | \mu_k, \Sigma_k),
    \label{eq:GMC}
    \vspace{-0.2cm}
\end{equation}
where $\theta = \{\lambda_k, \mu_k, \Sigma_k\}_{k=1}^{K}$ represents the parameters of all Gaussian distributions and $\lambda_k$ denotes the weight of each Gaussian distribution, with values between 0 and 1. We optimize the parameters $\theta$ using the Expectation-Maximization (EM) algorithm to maximize the log-likelihood function. Detailed algorithm steps are provided in Algorithm 2 in the appendix.

% \vspace{-0.2cm}
\textbf{Domain Feature Resampler.}
We use GMC to compress the feature set $\mathbf{X}^{l}_{\tau}$ into a set of parameters $\theta^{l}_{\tau} = \{\lambda^{l}_{\tau,k}, \mu^{l}_{\tau,k}, \Sigma^{l}_{\tau,k}\}_{k=1}^{K}$. However, using only the means $\mu^{l}_{\tau,k}$ for model training could reduce the model's generalization capability. Therefore, we use a Domain Feature Resampler (DFR) to resample $N_t$ pseudo-domain features to simulate the original feature distribution, where $N_t$ is the number of images in the $t$ domain. First, we sample $N_t$ Gaussian distribution indices based on their weights $(\lambda^{l}_{\tau,1}, \lambda^{l}_{\tau,2}, \dots, \lambda^{l}_{\tau,K})$, denoted as $\{k_i\}_{i=1}^{N_t}$. Then, we sample $\tilde{\mathbf{x}}_{i}$ from the $k_i$-th Gaussian distribution as follows:
% \vspace{-0.2cm}
\begin{equation}
    \tilde{\mathbf{x}}^{l}_{\tau,i} \sim \mathcal{N}(\mu^{l}_{\tau,k_i}, \Sigma^{l}_{\tau,k_i}), \quad \forall i \in \{1, 2, \dots, N_t\}.
    \label{eq:DFR}
    % \vspace{-0.2cm}
\end{equation}
Now we sample $N_t$ pseudo-domain features, denoted as $\tilde{\mathbf{X}}^{l}_{\tau} = \{\tilde{\mathbf{x}}^{l}_{\tau,i}\}_{i=1}^{N_t} \in \mathbb{R}^{N_t \times d}$.

% \vspace{-0.2cm}
\textbf{Training MDFN with GMC and DFR.}
We apply the GMC and DFR modules to process features of the $\frac{L}{2}$-th and $L$-th layers for each previous domain, \emph{i.e.}, from the first domain to the $(t-1)$-th domain. The feature $\mathbf{X}_{\tau}^{l}$ is compressed into $\theta_{\tau}^{l}$ and then resampled into $\tilde{\mathbf{X}}_{\tau}^{l}$, as formulated below:
\begin{equation}
\footnotesize
    \begin{aligned}
    \left\{
    \begin{array}{l}
        \theta_{\tau}^{l} = \text{GMC}(\mathbf{X}_{\tau}^{l},K), \\
        \tilde{\mathbf{X}}_{\tau}^{l} = \text{DFR}(\theta_{\tau}^{l},N_t),
    \end{array}
    \right.
    \forall \tau \in \{1,2,\dots,t-1\}, l \in \{\frac{L}{2}, L\},
    \end{aligned}
\end{equation}
where $K$ is a hyperparameter representing the number of Gaussian distributions.

When training on the $t$-th domain, we obtain the set $\{\tilde{\mathbf{X}}_1^l, \tilde{\mathbf{X}}_2^l, \cdots, \tilde{\mathbf{X}}_{t-1}^l, \mathbf{X}_{t}^l\}, \forall l \in \{\frac{L}{2}, L\}$, which contains a total of $2 \times t \times N_t$ features. Following \cref{sec:MDFN} and \cref{eq:MDFN}, we calculate the fused domain features $\tilde{\mathbf{X}}^{D}=\{\tilde{\mathbf{X}}_1^D, \tilde{\mathbf{X}}_2^D, \cdots, \tilde{\mathbf{X}}_{t-1}^D, \mathbf{X}_{t}^D\} \in \mathbb{R}^{(t \times N_{t}) \times d}$. We then randomly sample $\tilde{\mathbf{x}}^{D}_{\tau}$ from $\tilde{\mathbf{X}}^{D}$ to train the MDFN parameters $\Delta$, where the corresponding domain label is $\tau$. In conclusion, we address the class imbalance issue in MDFN training by preserving the key domain feature from previous domains. Additionally, we reduce the storage space required for saving domain features with the GMC and the DFR.

\begin{table*}[ht]
\footnotesize
\centering
\setlength{\tabcolsep}{6pt}

\begin{tabular}{c|l|c|ccc|ccc|c}
\hline
                           & \multicolumn{1}{c|}{}                          &                                              & \multicolumn{3}{c|}{DomainNet}                                                                                           & \multicolumn{3}{c|}{CDDB-Hard}                                                                                           & CORe50                                 \\ \cline{4-10} 
\multirow{-2}{*}{Backbone} & \multicolumn{1}{c|}{\multirow{-2}{*}{Methods}} & \multirow{-2}{*}{Buffer Size ($\downarrow$)} & $A_{T}(\uparrow)$                      & $F_{T}(\uparrow)$  & $S_{T}(\uparrow)$ & $A_{T}(\uparrow)$  & $F_{T}(\uparrow)$  & $S_{T}(\uparrow)$ & $A_{T}(\uparrow)$  \\ \hline
                           & DyTox~\cite{douillard2022dytox}                                          & 50/class                                     & 62.94                                  & -                                      & N/A                                    & 86.21                                  & -1.55                                  & N/A                                    & 79.21                                  \\ \cline{2-10} 
                           & EWC~\cite{kirkpatrick2017overcoming}                                            &                                              & 47.62                                  & -                                      & N/A                                    & 50.59                                  & -42.62                                 & N/A                                    & 74.82                                  \\
                           & LwF~\cite{li2017learning}                                            &                                              & 49.19                                  & -5.01                                  & N/A                                    & 60.94                                  & -13.53                                 & N/A                                    & 75.45                                  \\
                           & L2P~\cite{l2p}                                            &                                              & 40.15                                  & -2.25                                  & N/A                                    & 61.28                                  & -9.23                                  & N/A                                    & 78.33                                  \\
                           & DualPrompt~\cite{dualprompt}                                     &                                              & 43.79                                  & -2.03                                  & N/A                                    & 64.80                                  & -8.74                                  & N/A                                    & 80.25                                  \\
                           & CODA-P~\cite{coda}                                         &                                              & 47.42                                  & -3.46                                  & N/A                                    & 70.54                                  & -5.53                                  & N/A                                    & 85.68                                  \\
                           & PINA~\cite{PINA}                                           &                                              & 54.86                                  & -2.24                                  & 80.55                                  & 77.35                                  & -0.98                                  & 82.27                                  & 86.74                                  \\
                           & C-Prompt~\cite{cprompt}                                       &                                              & 58.68                                  & -1.34                                  & N/A                                    & -                                      & -                                      & N/A                                    & 85.31                                  \\ \cline{2-2} \cline{4-10} 
                           & S-iPrompts~\cite{sprompt}                                     &                                              & 50.62                                  & -2.85                                  & 80.35                                  & 74.51                                  & -1.30                                  & 81.79                                  & 83.13                                  \\
                           & \cellcolor[HTML]{E0E0E0}S-iPrompts+SOYO        &                                              & \cellcolor[HTML]{E0E0E0}53.32          & \cellcolor[HTML]{E0E0E0}-1.94          & \cellcolor[HTML]{E0E0E0}85.37          & \cellcolor[HTML]{E0E0E0}75.73          & \cellcolor[HTML]{E0E0E0}-0.80          & \cellcolor[HTML]{E0E0E0}89.84          & \cellcolor[HTML]{E0E0E0}83.97          \\
                           & \cellcolor[HTML]{E0E0E0}S-iPrompts+oracle      &                                              & \cellcolor[HTML]{E0E0E0}58.29          & \cellcolor[HTML]{E0E0E0}0              & \cellcolor[HTML]{E0E0E0}100            & \cellcolor[HTML]{E0E0E0}77.01          & \cellcolor[HTML]{E0E0E0}0              & \cellcolor[HTML]{E0E0E0}100            & \cellcolor[HTML]{E0E0E0}85.42          \\
                           & PINA-D~\cite{PINA}                                         &                                              & 62.23                                  & -1.83                                  & 80.55                                  & 78.20                                  & -0.73                                  & 82.27                                  & 90.01                                  \\
                           & \cellcolor[HTML]{E0E0E0}\textbf{PINA-D+SOYO}   &                                              & \cellcolor[HTML]{E0E0E0}\textbf{65.25} & \cellcolor[HTML]{E0E0E0}\textbf{-1.26} & \cellcolor[HTML]{E0E0E0}\textbf{85.37} & \cellcolor[HTML]{E0E0E0}\textbf{80.35} & \cellcolor[HTML]{E0E0E0}\textbf{-0.39} & \cellcolor[HTML]{E0E0E0}\textbf{89.84} & \cellcolor[HTML]{E0E0E0}\textbf{90.77} \\
\multirow{-14}{*}{ViT-B~\cite{vit}}   & \cellcolor[HTML]{E0E0E0}PINA-D+oracle          & \multirow{-13}{*}{0/class}                   & \cellcolor[HTML]{E0E0E0}70.56          & \cellcolor[HTML]{E0E0E0}0              & \cellcolor[HTML]{E0E0E0}100            & \cellcolor[HTML]{E0E0E0}81.22          & \cellcolor[HTML]{E0E0E0}0              & \cellcolor[HTML]{E0E0E0}100            & \cellcolor[HTML]{E0E0E0}91.30          \\ \hline \hline
                           & PINA~\cite{PINA}                                           &                                              & 69.06                                  & -1.59                                  & 83.91                                  & 85.71                                  & -0.51                                  & 81.44                                  & 87.38                                  \\
                           & MoP-CLIP~\cite{nicolas2024mop}                                       &                                              & 69.70                                  & -                                      & -                                      & 88.54                                  & -0.79                                  & -                                      & \textbf{92.29}                         \\ \cline{2-2} \cline{4-10} 
                           & S-liPrompts~\cite{sprompt}                                    &                                              & 67.78                                  & -1.64                                  & 84.05                                  & 88.65                                  & -0.69                                  & 80.13                                  & 89.06                                  \\
                           & \cellcolor[HTML]{E0E0E0}S-iPrompts+SOYO        &                                              & \cellcolor[HTML]{E0E0E0}69.20          & \cellcolor[HTML]{E0E0E0}-1.17          & \cellcolor[HTML]{E0E0E0}88.16          & \cellcolor[HTML]{E0E0E0}89.31          & \cellcolor[HTML]{E0E0E0}-0.48          & \cellcolor[HTML]{E0E0E0}89.57          & \cellcolor[HTML]{E0E0E0}89.41          \\
                           & \cellcolor[HTML]{E0E0E0}S-iPrompts+oracle      &                                              & \cellcolor[HTML]{E0E0E0}70.62          & \cellcolor[HTML]{E0E0E0}0              & \cellcolor[HTML]{E0E0E0}100            & \cellcolor[HTML]{E0E0E0}90.26          & \cellcolor[HTML]{E0E0E0}0              & \cellcolor[HTML]{E0E0E0}100            & \cellcolor[HTML]{E0E0E0}90.31          \\
                           & PINA-D~\cite{PINA}                                         &                                              & 73.27                                  & -1.12                                  & 83.91                                  & 89.03                                  & -0.32                                  & 81.44                                  & 89.23                                  \\
                           & \cellcolor[HTML]{E0E0E0}\textbf{PINA-D+SOYO}   &                                              & \cellcolor[HTML]{E0E0E0}\textbf{74.72} & \cellcolor[HTML]{E0E0E0}\textbf{-0.83} & \cellcolor[HTML]{E0E0E0}\textbf{88.16} & \cellcolor[HTML]{E0E0E0}\textbf{90.12} & \cellcolor[HTML]{E0E0E0}\textbf{-0.20} & \cellcolor[HTML]{E0E0E0}\textbf{89.57} & \cellcolor[HTML]{E0E0E0}89.82          \\
\multirow{-8}{*}{CLIP~\cite{clip}}     & \cellcolor[HTML]{E0E0E0}PINA-D+oracle          & \multirow{-8}{*}{0/class}                    & \cellcolor[HTML]{E0E0E0}76.86          & \cellcolor[HTML]{E0E0E0}0              & \cellcolor[HTML]{E0E0E0}100            & \cellcolor[HTML]{E0E0E0}91.85          & \cellcolor[HTML]{E0E0E0}0              & \cellcolor[HTML]{E0E0E0}100            & \cellcolor[HTML]{E0E0E0}90.56          \\ \hline
\end{tabular}
% \vspace{-0.1cm}
\caption{\textbf{Comparison results for the DIC task.}
% $A_{T}$, $F_{T}$, $S_{T}$ represent the average accuracy, the forgetting degree, and the domain selection accuracy after training on all $T$ tasks, respectively.
Note that the training and test sets in the CORe50 dataset come from non-overlapping domains; therefore, $F_{T}$ and $S_{T}$ are not applicable to the CORe50 dataset. \textbf{Bold} denotes the best parameter-isolation DIC results. ``Oracle'' refers to an upper bound of the baseline, assuming that the domain label of the test sample is known.}
\label{tab:DIC}
% \vspace{-0.1cm}
\end{table*}

\begin{table*}[ht]
\footnotesize
\centering
\setlength{\tabcolsep}{2pt}

\begin{tabular}{l|cccccc|ccccc}
\hline
\multicolumn{1}{c|}{}                          & \multicolumn{6}{c|}{Pascal VOC series}                                                                                                                                                                                                                   & \multicolumn{5}{c}{BDD100K series}                                                                                                                                                                               \\ \cline{2-12} 
\multicolumn{1}{c|}{\multirow{-2}{*}{Methods}} & \multicolumn{1}{c|}{Buffer Size ($\downarrow$)}  & Session 1                             & Session 2                             & Session 3                             & Session 4                             & $S_{T}(\uparrow)$                     & \multicolumn{1}{c|}{Buffer Size}                 & Session 1                             & Session 2                             & Session 3                             & $S_{T}(\uparrow)$                     \\ \hline
Upper-bound                                    & \multicolumn{1}{c|}{-}                           & 86.4                                  & 72.6                                  & 69.4                                  & 67.6                                  & N/A                                   & \multicolumn{1}{c|}{-}                           & 52.1                                  & 57.0                                  & 58.5                                  & N/A                                   \\ \hline
TP-DIOD-B                                      & \multicolumn{1}{c|}{150/domain}                  & 86.4                                  & 65.8                                  & 62.1                                  & 57.5                                  & N/A                                   & \multicolumn{1}{c|}{200/domain}                  & 52.1                                  & 53.4                                  & 51.5                                  & N/A                                   \\ \hline
FT-seq                                         & \multicolumn{1}{c|}{}                            & 86.4                                  & 57.5                                  & 52.6                                  & 49.5                                  & N/A                                   & \multicolumn{1}{c|}{}                            & 52.1                                  & 51.6                                  & 43.6                                  & N/A                                   \\
FT-FC                                          & \multicolumn{1}{c|}{}                            & 86.4                                  & 59.1                                  & 54.4                                  & 44.2                                  & N/A                                   & \multicolumn{1}{c|}{}                            & 52.1                                  & 51.3                                  & 48.0                                  & N/A                                   \\
MCC~\cite{mcc}                                            & \multicolumn{1}{c|}{}                            & 86.4                                  & 47.6                                  & 34.4                                  & 23.7                                  & N/A                                   & \multicolumn{1}{c|}{}                            & 52.1                                  & 44.3                                  & 36.1                                  & N/A                                   \\
IRG~\cite{irg}                                            & \multicolumn{1}{c|}{}                            & 86.4                                  & 51.5                                  & 43.7                                  & 33.2                                  & N/A                                   & \multicolumn{1}{c|}{}                            & 52.1                                  & 49.3                                  & 38.7                                  & N/A                                   \\
LwF~\cite{li2017learning}                                            & \multicolumn{1}{c|}{}                            & 86.4                                  & 60.4                                  & 53.6                                  & 53.2                                  & N/A                                   & \multicolumn{1}{c|}{}                            & 52.1                                  & 52.1                                  & 44.1                                  & N/A                                   \\
PASS~\cite{pass}                                           & \multicolumn{1}{c|}{}                            & 86.4                                  & 61.7                                  & 51.4                                  & 49.8                                  & N/A                                   & \multicolumn{1}{c|}{}                            & 52.1                                  & 51.7                                  & 43.3                                  & N/A                                   \\
L2P~\cite{l2p}                                            & \multicolumn{1}{c|}{}                            & 86.4                                  & 59.9                                  & 55.2                                  & 45.5                                  & N/A                                   & \multicolumn{1}{c|}{}                            & 52.1                                  & 51.5                                  & 47.7                                  & N/A                                   \\
S-Prompts~\cite{sprompt}                                       & \multicolumn{1}{c|}{}                            & 86.4                                  & 59.4                                  & 54.3                                  & 45.0                                  & 71.1                                  & \multicolumn{1}{c|}{}                            & 52.1                                  & 51.6                                  & 49.4                                  & 93.7                                  \\
CIFRCN~\cite{cifrcn}                                         & \multicolumn{1}{c|}{}                            & 86.4                                  & 65.3                                  & 57.7                                  & 53.5                                  & N/A                                   & \multicolumn{1}{c|}{}                            & 52.1                                  & 51.8                                  & 48.9                                  & N/A                                   \\
ERD~\cite{erd}                                            & \multicolumn{1}{c|}{}                            & 86.4                                  & 58.9                                  & 50.7                                  & 48.7                                  & N/A                                   & \multicolumn{1}{c|}{}                            & 52.1                                  & 51.1                                  & 48.1                                  & N/A                                   \\ \cline{1-1} \cline{3-7} \cline{9-12} 
LDB (K\&K)                                     & \multicolumn{1}{c|}{}                            & 86.4                                  & 67.4                                  & 62.9                                  & 55.9                                  & 71.1                                  & \multicolumn{1}{c|}{}                            & 52.1                                  & 52.1                                  & 50.5                                  & 93.7                                  \\
LDB (NMC)                                      & \multicolumn{1}{c|}{}                            & 86.4                                  & 68.1                                  & 64.2                                  & 56.8                                  & 77.1                                  & \multicolumn{1}{c|}{}                            & 52.1                                  & 52.3                                  & 51.1                                  & 96.4                                  \\
\cellcolor[HTML]{E0E0E0}\textbf{LDB+SOYO}      & \multicolumn{1}{c|}{}                            & \cellcolor[HTML]{E0E0E0}\textbf{86.4} & \cellcolor[HTML]{E0E0E0}\textbf{69.2} & \cellcolor[HTML]{E0E0E0}\textbf{65.3} & \cellcolor[HTML]{E0E0E0}\textbf{59.6} & \cellcolor[HTML]{E0E0E0}\textbf{96.7} & \multicolumn{1}{c|}{}                            & \cellcolor[HTML]{E0E0E0}\textbf{52.1} & \cellcolor[HTML]{E0E0E0}\textbf{52.4} & \cellcolor[HTML]{E0E0E0}\textbf{51.7} & \cellcolor[HTML]{E0E0E0}\textbf{98.2} \\
\cellcolor[HTML]{E0E0E0}LDB+oracle             & \multicolumn{1}{c|}{\multirow{-14}{*}{0/domain}} & \cellcolor[HTML]{E0E0E0}86.4          & \cellcolor[HTML]{E0E0E0}69.6          & \cellcolor[HTML]{E0E0E0}65.9          & \cellcolor[HTML]{E0E0E0}59.9          & \cellcolor[HTML]{E0E0E0}100           & \multicolumn{1}{c|}{\multirow{-14}{*}{0/domain}} & \cellcolor[HTML]{E0E0E0}52.1          & \cellcolor[HTML]{E0E0E0}52.4          & \cellcolor[HTML]{E0E0E0}52.0          & \cellcolor[HTML]{E0E0E0}100           \\ \hline
\end{tabular}
% \vspace{-0.1cm}
\caption{\textbf{Comparison results of mAP for the DIOD task.} K\&K represents the KMeans and K-Nearest Neighbors algorithms, while NMC refers to the nearest mean classifier. Both are used to predict the domain of the test samples. \textbf{Bold} denotes the best parameter-isolation DIOD results. ``Oracle'' refers to an upper bound of the baseline, assuming that the domain label of the test sample is known.}
\label{tab:DIOD}
\vspace{-0.2cm}
\end{table*}

\begin{table}[ht]
\footnotesize
\centering
\setlength{\tabcolsep}{3pt}

\begin{tabular}{c|l|cccc|c}
\hline
Metrics                                                                            & Methods                                   & Alarm                                  & Cough                                  & DD                                     & MG                                     & Avg.                                   \\ \hline
                                                                                   & Unprocessed                               & 2.98                                   & 2.94                                   & 2.37                                   & 2.37                                   & 2.67                                   \\
                                                                                   & Pre-trained                               & 11.36                                  & 7.10                                   & 13.70                                  & 14.00                                  & 11.54                                  \\
                                                                                   & FT-seq                                    & 3.00                                   & 3.19                                   & 6.22                                   & 25.82                                  & 9.56                                   \\
                                                                                   & SERIL~\cite{seril}                                     & 4.82                                   & 4.09                                   & 10.76                                  & 23.94                                  & 10.90                                  \\
                                                                                   & LNA~\cite{LNA}                                       & 18.07                                  & 16.83                                  & 14.85                                  & 14.52                                  & 16.07                                  \\
                                                                                   & \cellcolor[HTML]{E0E0E0}\textbf{LNA+SOYO} & \cellcolor[HTML]{E0E0E0}\textbf{18.50} & \cellcolor[HTML]{E0E0E0}\textbf{17.86} & \cellcolor[HTML]{E0E0E0}\textbf{15.35} & \cellcolor[HTML]{E0E0E0}\textbf{18.27} & \cellcolor[HTML]{E0E0E0}\textbf{17.50} \\
\multirow{-7}{*}{\begin{tabular}[c]{@{}c@{}}SI-SNR\\ (dB, $\uparrow$)\end{tabular}} & \cellcolor[HTML]{E0E0E0}LNA+oracle        & \cellcolor[HTML]{E0E0E0}18.72          & \cellcolor[HTML]{E0E0E0}18.13          & \cellcolor[HTML]{E0E0E0}15.69          & \cellcolor[HTML]{E0E0E0}21.54          & \cellcolor[HTML]{E0E0E0}18.52          \\ \hline
                                                                                   & Unprocessed                               & 3.05                                   & 3.00                                   & 2.45                                   & 2.45                                   & 2.74                                   \\
                                                                                   & Pre-trained                               & 11.55                                  & 7.18                                   & 14.07                                  & 14.45                                  & 11.81                                  \\
                                                                                   & FT-seq                                    & 3.07                                   & 3.25                                   & 6.31                                   & 26.14                                  & 9.69                                   \\
                                                                                   & SERIL~\cite{seril}                                     & 4.90                                   & 4.15                                   & 10.93                                  & 23.60                                  & 10.90                                  \\
                                                                                   & LNA~\cite{LNA}                                       & 18.47                                  & 17.13                                  & 15.27                                  & 14.75                                  & 16.41                                  \\
                                                                                   & \cellcolor[HTML]{E0E0E0}\textbf{LNA+SOYO} & \cellcolor[HTML]{E0E0E0}\textbf{18.92} & \cellcolor[HTML]{E0E0E0}\textbf{18.19} & \cellcolor[HTML]{E0E0E0}\textbf{15.79} & \cellcolor[HTML]{E0E0E0}\textbf{18.57} & \cellcolor[HTML]{E0E0E0}\textbf{17.87} \\
\multirow{-7}{*}{\begin{tabular}[c]{@{}c@{}}SDR\\ (dB, $\uparrow$)\end{tabular}}    & \cellcolor[HTML]{E0E0E0}LNA+oracle        & \cellcolor[HTML]{E0E0E0}19.13          & \cellcolor[HTML]{E0E0E0}18.46          & \cellcolor[HTML]{E0E0E0}16.10          & \cellcolor[HTML]{E0E0E0}21.90          & \cellcolor[HTML]{E0E0E0}18.90          \\ \hline
                                                                                   & Unprocessed                               & 1.79                                   & 1.75                                   & 1.78                                   & 2.36                                   & 1.92                                   \\
                                                                                   & Pre-trained                               & 3.09                                   & 2.15                                   & 2.66                                   & 2.89                                   & 2.70                                   \\
                                                                                   & FT-seq                                    & 1.79                                   & 1.76                                   & 1.89                                   & 4.21                                   & 2.41                                   \\
                                                                                   & SERIL~\cite{seril}                                     & 1.88                                   & 1.83                                   & 2.14                                   & 4.09                                   & 2.49                                   \\
                                                                                   & LNA~\cite{LNA}                                       & 3.32                                   & 3.09                                   & 2.86                                   & 3.26                                   & 3.13                                   \\
                                                                                   & \cellcolor[HTML]{E0E0E0}\textbf{LNA+SOYO} & \cellcolor[HTML]{E0E0E0}\textbf{3.40}  & \cellcolor[HTML]{E0E0E0}\textbf{3.23}  & \cellcolor[HTML]{E0E0E0}\textbf{2.93}  & \cellcolor[HTML]{E0E0E0}\textbf{3.61}  & \cellcolor[HTML]{E0E0E0}\textbf{3.29}  \\
\multirow{-7}{*}{PESQ ($\uparrow$)}                                                 & \cellcolor[HTML]{E0E0E0}LNA+oracle        & \cellcolor[HTML]{E0E0E0}3.44           & \cellcolor[HTML]{E0E0E0}3.27           & \cellcolor[HTML]{E0E0E0}3.00           & \cellcolor[HTML]{E0E0E0}3.93           & \cellcolor[HTML]{E0E0E0}3.41           \\ \hline
\end{tabular}
% \vspace{-0cm}
\caption{\textbf{Comparison results on the WSJ0 synthetic dataset for the DISE task.} \textbf{Bold} denotes the best parameter-isolation DISE results. ``Oracle'' refers to an upper bound of the baseline, assuming that the domain label of the test sample is known.}
\label{tab:DISE}
% \vspace{-0.1cm}
\end{table}

\begin{table}[ht]
\footnotesize
\centering
\setlength{\tabcolsep}{2.7pt}

\begin{tabular}{l|c|c|cccc}
\hline
Methods                     & Session                      & Domain & $S'_{1}$                               & $S'_{2}$                               & $S'_{3}$                               & $S'_{4}$                               \\ \hline
                            &                              & Alarm  & \cellcolor[HTML]{E0E0E0}78.80          & 21.20                                  & 0.00                                   & 0.00                                   \\
                            &                              & Cough  & 9.22                                   & \cellcolor[HTML]{E0E0E0}86.33          & 2.00                                   & 2.45                                   \\
                            &                              & DD     & 0.31                                   & 6.14                                   & \cellcolor[HTML]{E0E0E0}77.11          & 16.44                                  \\
\multirow{-4}{*}{LNA+K\&K}  & \multirow{-4}{*}{Session 4*}  & MG     & 0.15                                   & 3.99                                   & 39.32                                  & \cellcolor[HTML]{E0E0E0}56.53          \\ \hline
                            & Session 1                   & Alarm  & \cellcolor[HTML]{E0E0E0}100.00         &                                        &                                        &                                        \\ \cdashline{2-7} 
                            &                              & Alarm  & \cellcolor[HTML]{E0E0E0}94.32          & 5.68                                   &                                        &                                        \\
                            & \multirow{-2}{*}{Session 2} & Cough  & 2.47                                   & \cellcolor[HTML]{E0E0E0}97.53          &                                        &                                        \\ \cdashline{2-7} 
                            &                              & Alarm  & \cellcolor[HTML]{E0E0E0}93.09          & 6.91                                   & 0.00                                   &                                        \\
                            &                              & Cough  & 3.84                                   & \cellcolor[HTML]{E0E0E0}95.70          & 0.46                                   &                                        \\
                            & \multirow{-3}{*}{Session 3} & DD     & 0.67                                   & 4.15                                   & \cellcolor[HTML]{E0E0E0}95.18          &                                        \\ \cdashline{2-7} 
                            &                              & Alarm  & \cellcolor[HTML]{E0E0E0}\textbf{90.32} & 9.68                                   & 0.00                                   & 0.00                                   \\
                            &                              & Cough  & 2.46                                   & \cellcolor[HTML]{E0E0E0}\textbf{96.77} & 0.31                                   & 0.46                                   \\
                            &                              & DD     & 0.77                                   & 2.30                                   & \cellcolor[HTML]{E0E0E0}\textbf{90.78} & 6.14                                   \\
\multirow{-10}{*}{LNA+SOYO} & \multirow{-4}{*}{Session 4}  & MG     & 1.08                                   & 1.99                                   & 19.20                                  & \cellcolor[HTML]{E0E0E0}\textbf{77.73} \\ \hline
\end{tabular}
% \vspace{-0cm}
\caption{\textbf{Parameter selection accuracy on the WSJ0 synthetic dataset.} $S'_{1}-S'_{4}$ represents the proportion of test samples that are classified as the Alarm, Cough, DD, or MG domains, respectively. Shaded values closer to 100 indicate higher accuracy in domain selection. * denotes that the LNA method only reports accuracy for Session 4. For reference, we also present the parameter selection accuracy of our method across Sessions 1-3.}
\label{tab:DISE2}
\vspace{-0.2cm}
\end{table}

\section{Experiments}
\subsection{Benchmark and Implementation}
\textbf{Tasks.} We evaluated our method on three tasks: Domain Incremental Image Classification (DIC), Object Detection (DIOD), and Speech Enhancement (DISE). These tasks cover two modalities (images and audio), demonstrating the versatility of our approach.

\textbf{Datasets.} We conducted extensive experiments on six datasets. \textbf{For DIC}, we used three datasets: DomainNet~\cite{peng2019moment}, CDDB~\cite{li2023continual}, and CORe50~\cite{lomonaco2017core50}. Specifically, \textbf{DomainNet} is a large-scale dataset containing over 586,000 images divided into 6 domains and 345 categories. \textbf{CDDB} is a dataset that includes multiple deepfake techniques for continual deepfake detection. We selected the hard track following~\cite{sprompt}. \textbf{CORe50} is an object recognition dataset comprising 50 categories and 11 domains, with each domain containing 15,000 images.

\textbf{DIOD} includes two datasets: Pascal VOC series and BDD100K series. The \textbf{Pascal VOC series} is an object detection dataset containing 4 domains (Pascal VOC 2007~\cite{everingham2010pascal}, Clipart, Watercolor, and Comic) and 6 categories. We followed the setup in~\cite{LDB}, using 5,923 images for training and 5,879 for testing. The \textbf{BDD100K series} is an autonomous driving dataset with 3 domains (BDD100K~\cite{yu2020bdd100k}, Cityscape~\cite{cordts2016cityscapes}, Rainy Cityscape~\cite{hu2019depth}) and 8 categories. We trained on 82,407 images and tested on 11,688 images.

\textbf{For DISE}, we validated our method on a synthetic dataset based on \textbf{WSJ0~\cite{varga1993assessment} and NOISEX-92~\cite{garofolo1993csr}}. Following~\cite{LNA}, we constructed a dataset containing 14 domains, with 10 types of noise for base model training, and 4 noises: Alarm, Cough, Destroyerops (DD), and MachineGun (MG) for incremental learning. The goal of DISE is to remove the noise and restore the original clean speech.

\textbf{Evaluation Metrics.}
We report the average classification accuracy ($A_T$), the forgetting degree ($F_T$), and the parameter selection accuracy ($S_T$) for DIC. Further details are in the appendix.
We evaluate the average precision (AP) with an IoU threshold of 0.5 and report the mean average precision (mAP) following LDB~\cite{LDB} for DIOD. 
For the DISE task, we measure three metrics following LNA~\cite{LNA}: SI-SNR, signal-to-distortion ratio (SDR), and perceptual evaluation of speech quality (PESQ) for DISE.

\textbf{Baselines.} 
We compare our approach with parameter-isolation methods, including S-Prompts~\cite{sprompt} and PINA~\cite{PINA} in the DIC task.
For DIOD, we use the LDB method~\cite{LDB} as the baseline to validate the effectiveness of our SOYO approach.
For DISE, we evaluate our method on top of the LNA~\cite{LNA}, a DISE method based on parameter isolation.

\textbf{Implementation Details.} 
We set the GMC hyperparameter $K$ to 2, 3, and 1 for DIC, DIOD, and DISE, respectively. $g_1$ and $g_2$ are designed as MLPs, where the input and output dimensions depend on the feature dimensions extracted by the backbone network, \emph{i.e.}, 768, 768, and 256 for DIC, DIOD, and DISE, respectively. The hidden layer dimension of the MLP is 16. $g_3$ is a linear layer used for classification, with an output dimension of $t$ for the $t$-th domain. The learning rate, weight decay, and number of training epochs are set to 0.01, $2 \times 10^{-4}$, and 100, respectively.

\subsection{Main Results}

\textbf{Results for DIC.} 
\cref{tab:DIC} presents the experimental results on the DomainNet, CDDB-Hard, and CORe50 datasets for the DIC task. We evaluated our approach on two backbones and compared it with two representative parameter-isolation DIC methods: S-Prompts~\cite{sprompt} and PINA~\cite{PINA}. The results demonstrate that our SOYO method consistently improves parameter selection accuracy by 5-7\%, increasing average accuracy from 62.23\% to 65.25\% on DomainNet and from 78.20\% to 80.35\% on CDDB. For the CORe50 dataset, despite non-overlapping domains between the training set (8 domains) and the testing set (3 domains), our approach still achieved higher average accuracy than the S-Prompts and PINA. These improvements indicate that SOYO not only enhances domain prediction accuracy in seen domains but also improves generalization to unseen domains by better clustering the seen domains.

\textbf{Results for DIOD.} 
\cref{tab:DIOD} presents the results on the Pascal VOC series and BDD100K series datasets. We used LDB, a parameter-isolation DIOD method, as the baseline for the experiments. 
Our SOYO method improved parameter selection accuracy at most 19.6\% on the Pascal VOC series, significantly outperforming the NMC method (77.1\% $\rightarrow$ 96.7\%). Additionally, we reported the performance of the LDB method in the ideal ``oracle'' scenario, showing that our approach increased the mAP metric by 2.8\%, with only a 0.3\% difference compared to the ``oracle'' scenario.

\textbf{Results for DISE.} 
\cref{tab:DISE} reports the experimental results for the DISE task. We conducted evaluations across three metrics following the LNA method. The proposed method consistently outperformed LNA, with the SI-SNR metric improving from 16.07 to 17.50, the SDR metric from 16.41 to 17.87, and the PESQ metric from 3.13 to 3.29. Additionally, \cref{tab:DISE2} reports the parameter selection accuracy at each session stage. The results in session 4 show that our SOYO increased the accuracy by an average of 14.21\% (74.69\% $\rightarrow$ 88.90\%) over the KMeans \& KNN method used by LNA, thereby improving the overall performance consistently.

\subsection{Ablation Studies}

\begin{figure}[t]
  \centering
  \includegraphics[width=0.48\textwidth]{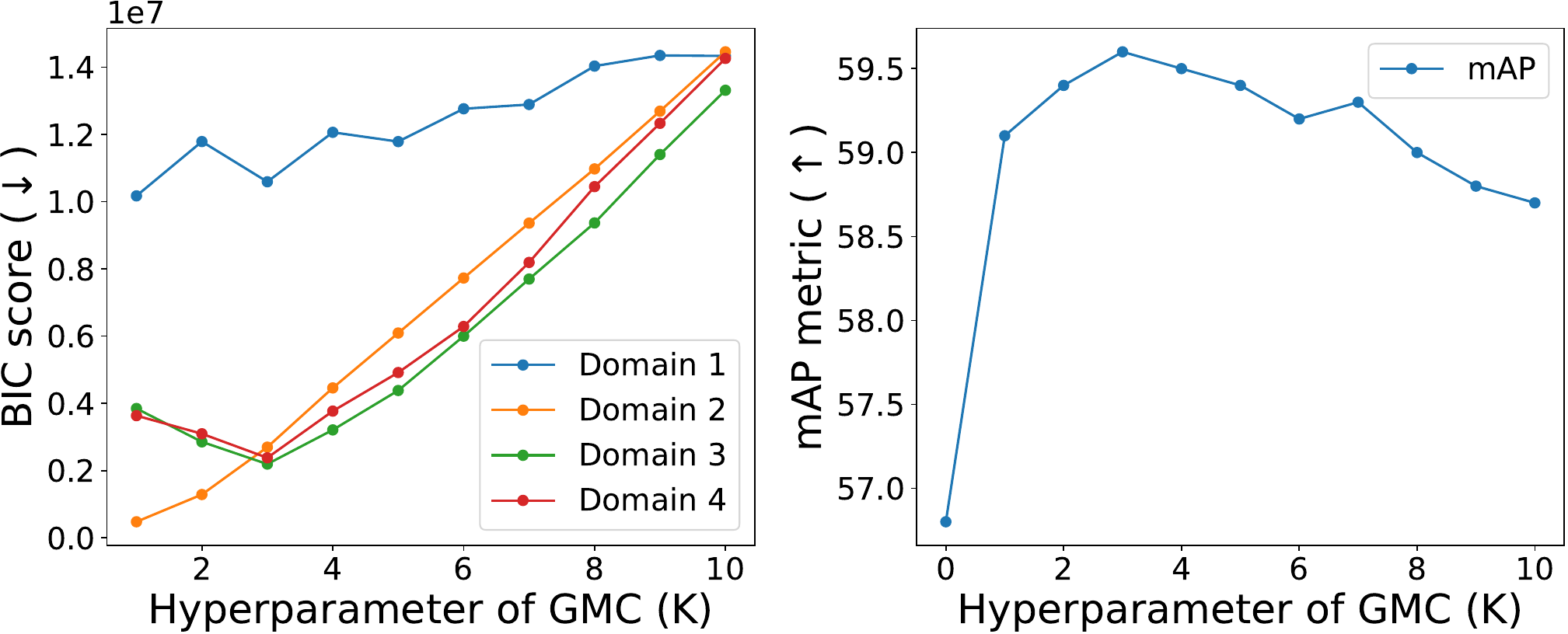}
  % \vspace{-0.4cm}
  \caption{\textbf{BIC score and mAP on the Pascal VOC series dataset.} In the right table, $K=0$ represents the baseline without GMC.}
  \label{fig:bic}
  % \vspace{-0cm}
\end{figure}

\begin{table}[t]
\footnotesize
\centering
\setlength{\tabcolsep}{2pt}
\begin{tabular}{l|cc|cccc}
\hline
Method                     & Memory     & \#Params            & $A_{T}$        & $\Delta A_{T}$ & $S_{T}$        & $\Delta S_{T}$ \\ \hline
KMeans\&KNN                & 0.004\%          & \multirow{3}{*}{0}      & 61.58          & -0.65          & 80.35          & -0.20          \\
Nearest Mean Classifier    & 0.001\%          &                         & 61.16          & -1.07          & 79.58          & -0.97          \\
Patch Shuffle Selector     & 0.001\%          &                         & 62.23          & 0              & 80.55          & 0              \\ \hline
SOYO (Mean\&std)           & 0.002\%          & \multirow{5}{*}{0.06\%} & 63.48          & \textbf{+1.25} & 82.83          & \textbf{+2.28} \\
SOYO (PCA, $N=10$)         & 0.009\%          &                         & 63.74          & \textbf{+1.51} & 83.15          & \textbf{+2.60} \\
SOYO (PCA, $N=100$)        & 0.085\%          &                         & 64.30          & \textbf{+2.07} & 83.88          & \textbf{+3.33} \\
SOYO (GMC, $K=1$)          & 0.647\%          &                         & 64.91          & \textbf{+2.68} & 84.96          & \textbf{+4.41} \\
\textbf{SOYO (GMC, $K=2$)} & \textbf{1.293\%} &                         & \textbf{65.25} & \textbf{+3.02} & \textbf{85.37} & \textbf{+4.82} \\ \hline
\end{tabular}
\caption{\textbf{Comparison of memory usage, extra parameters, and performance of different domain feature compression methods on the DomainNet dataset.} The baseline is PINA-D for the DIC task. ``Mean\&std'' indicates that only the means and standard deviations of the domain features are saved. ``PCA'' denotes Principal Component Analysis, and $N$ represents the number of principal components. The backbone model is ViT-B/16. The percentages indicate the ratio of memory usage or extra trainable parameters relative to the parameter count of the backbone model.}
\label{tab:memory}
\vspace{-0.2cm}
\end{table}

\textbf{Select the Hyperparameter $K$ for GMC.}
The hyperparameter $K$ represents the number of Gaussian components in the GMC. To select an appropriate value for $K$, we use the Bayesian Information Criterion (BIC), a model selection criterion that balances model complexity (\emph{i.e.}, the number of parameters) with the fit to the data. BIC penalizes models with more parameters to help prevent overfitting; therefore, a lower BIC value suggests a better balance between model simplicity and data fit. Using the Pascal VOC series dataset as an example, we evaluated BIC values for $K$ ranging from 1 to 10. As shown in \cref{fig:bic}, the results indicate that $K = 3$ is optimal for this dataset. We also tested the mean Average Precision (mAP) metric on the same setting, and the results similarly show that $K = 3$ achieves the best performance.

\textbf{Memory Usage and Extra Parameters.}
The proposed SOYO is a training-based approach, which requires additional trainable parameters and storage for previous domain features. In \cref{tab:memory}, we report SOYO's memory usage and final accuracy combined with different domain feature compression methods on the DomainNet dataset for the DIC task. Specifically, we experimented with three methods: Mean\&std, PCA, and GMC, and adjusted parameters for PCA and GMC to control memory usage. The results show that all methods under the SOYO framework outperform existing training-free approaches. With a maximum of only 1.293\% additional memory and 0.06\% extra trainable parameters, our approach achieves a 3.02\% improvement in accuracy over the PSS method.

\begin{table}[t]
\footnotesize
\centering
\setlength{\tabcolsep}{2.6pt}

\begin{tabular}{c|l|cc|cc}
\hline
\multirow{2}{*}{MDFN}      & \multirow{2}{*}{Feature Extraction Layers}  & \multicolumn{2}{c|}{DomainNet}  & \multicolumn{2}{c}{CDDB}        \\ \cline{3-6} 
                           &                          & $A_{T}$        & $S_{T}$        & $A_{T}$        & $S_{T}$        \\ \hline
\ding{55} & 12 (w/o MDFN)            & 63.61          & 82.82          & 79.36          & 86.18          \\
\ding{51} & 3+12                     & 64.93          & 85.01          & 80.11          & 89.21          \\
\ding{51} & \textbf{6+12 (Proposed)} & \textbf{65.25} & \textbf{85.37} & \textbf{80.35} & \textbf{89.84} \\
\ding{51} & 9+12                     & 64.68          & 84.67          & 79.94          & 88.86          \\
\ding{51} & 3+6+9+12                 & 65.12          & 85.20          & 80.26          & 89.54          \\ \hline
\end{tabular}
\vspace{-0.1cm}
\caption{\textbf{Ablation study of features used in MDFN for the DIC task.} Existing methods use features from the last layer of the transformer (the 12th layer in ViT-B) as domain features. We use this as our baseline.}
\label{tab:ab_MDFN}
\vspace{-0.2cm}
\end{table}

\textbf{Ablation for MDFN.}
We designed the MDFN to integrate shallow spatial information with global semantic features to obtain more discriminative domain features. To verify the effectiveness of MDFN, we conducted an ablation study as shown in \cref{tab:ab_MDFN}. The results demonstrate that incorporating shallow spatial features as auxiliary information enhances the discriminative ability of the domain features. Additionally, we experimented with the selection of layers. \cref{tab:ab_MDFN} indicate that fusing features from the 6th and 12th layers of ViT-B achieved the best results, improving domain prediction accuracy by approximately 3\% on the DomainNet and CDDB datasets compared to the baseline.

\subsection{Discussion and Visualization}

\begin{table}[t]
  \footnotesize
  \centering
  \setlength{\tabcolsep}{3pt}
  \begin{tabular}{lccccc}
    \hline
                  & Training Time  & GPU Memory & $A_{T}$\\
    \hline
    Baseline+\textcolor{cyan}{NMC}  & 18.3h+\textcolor{cyan}{0h}   & 16.4GB+\textcolor{cyan}{0GB} & 61.16\\
    Baseline+\textcolor{blue}{KNN}  & 18.3h+\textcolor{blue}{0h}   & 16.4GB+\textcolor{blue}{0GB} & 61.58\\
    \rowcolor[HTML]{E0E0E0}
    Baseline+\textcolor{red}{SOYO} (ours) & 18.3h+\textcolor{red}{1.6h}  & 16.4GB+\textcolor{red}{2.1GB} & \textcolor{red}{65.25}\\
    \hline
  \end{tabular}
  \vspace{-0.1cm}
  \caption{\textbf{Details of computational overheads.} NMC refers to the Nearest Mean Classifier, while KNN represents the K-Nearest Neighbors algorithm.}
  \label{tab:overhead}
  \vspace{-0.3cm}
\end{table}

\begin{figure}[t]
  \centering
  \includegraphics[width=0.45\textwidth]{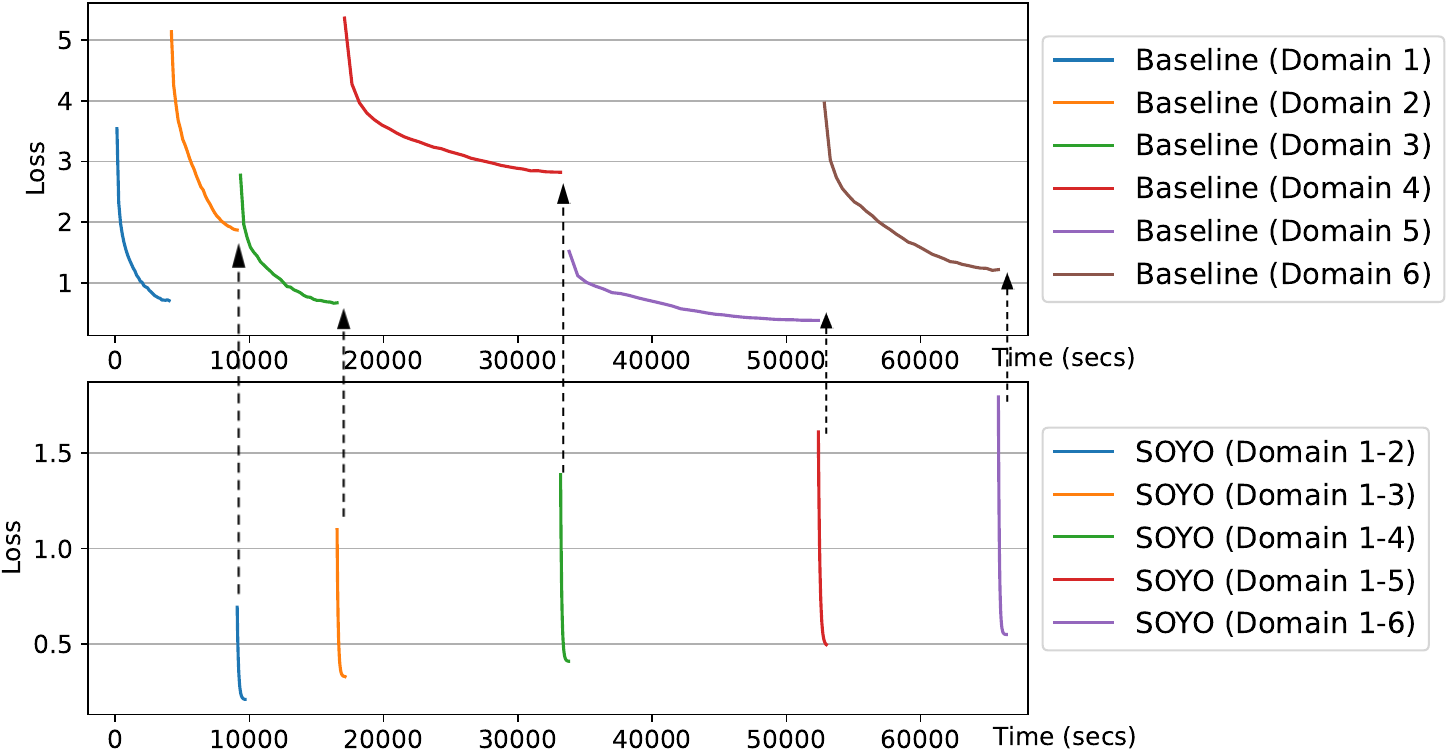}
  \vspace{-0.1cm}
  \caption{Model training convergence curves on DomainNet. The black dashed arrow indicates that SOYO training is conducted on each domain (starting from the second domain) following the baseline training.}
  \label{fig:curve}
  \vspace{-0.1cm}
\end{figure}

\textbf{Model Efficiency and Overheads.}
In \cref{tab:overhead}, we provided detailed analysis on training time, storage, GPU memory, and Accuracy of our SOYO. Compared to NMC or KNN methods, it incurs an additional 9\% training time and 13\% GPU memory, yielding a 4\% accuracy improvement on the DomainNet dataset. More details on training convergence curves are provided in \cref{fig:curve}.

\begin{figure}[t]
  \centering
  \includegraphics[width=0.45\textwidth]{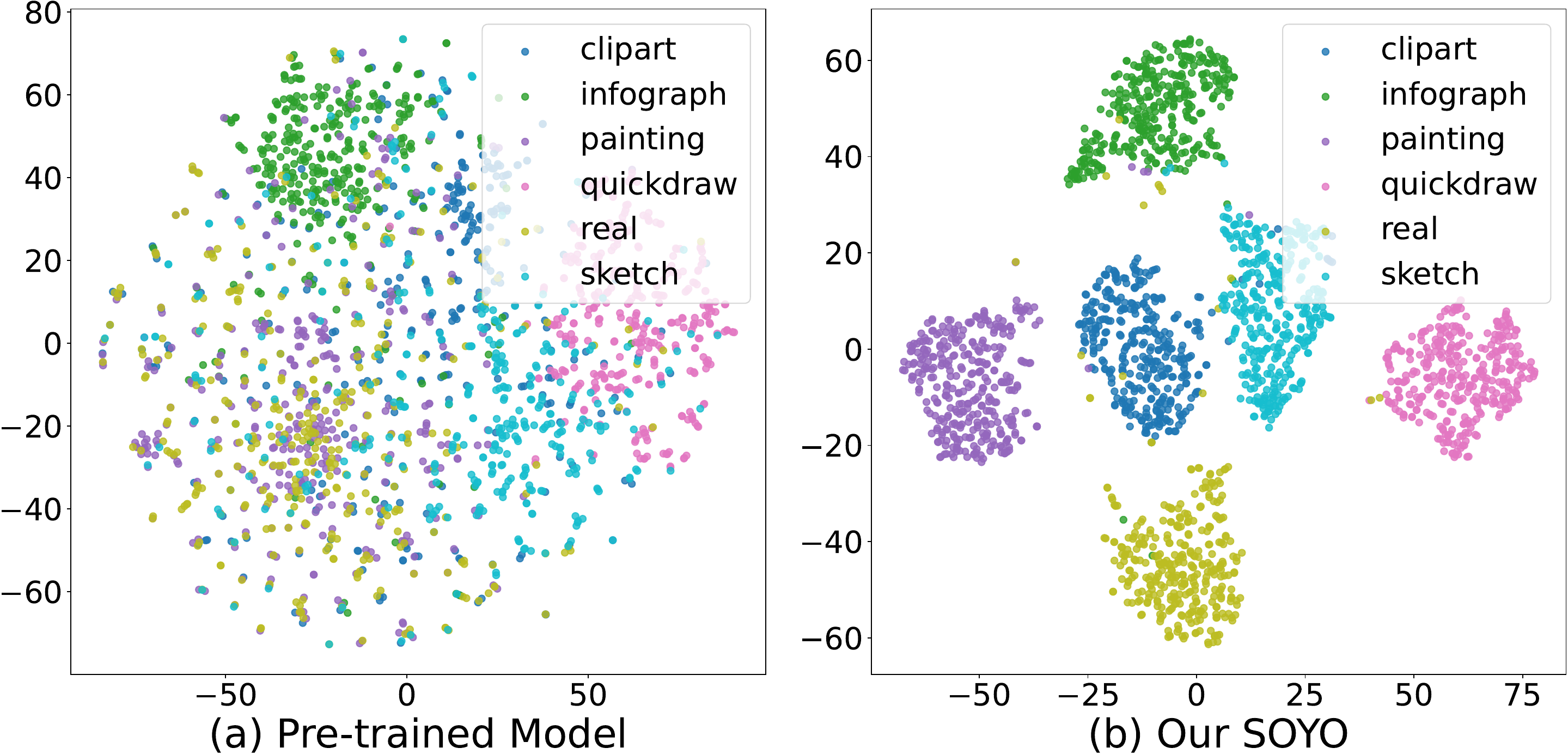}
  \vspace{-0.1cm}
  \caption{\textbf{t-SNE visualization of image features.} Features are extracted by randomly selecting one image from each domain and class (345 classes across 6 domains) in the DomainNet dataset. (a) visualized the features extracted from the pre-trained ViT-B, and (b) shows the features $x^{D}$ in the MDFN module (see \cref{fig:framework}).}
  \label{fig:tsne}
  \vspace{-0.2cm}
\end{figure}

\textbf{Comparison of Feature Space Distributions.}
\cref{fig:tsne} shows the t-SNE visualization of the pre-trained model and our SOYO method. Training-free methods rely on pre-trained features to compute domain cluster centers, so suffer from low domain selection accuracy. In contrast, the proposed SOYO method shows clearer domain feature separation, indicating superior performance in domain prediction and parameter selection.

\section{Conclusion}
In this paper, we introduced SOYO, a novel and lightweight framework that enhances parameter selection accuracy in parameter-isolation domain incremental learning. By incorporating the Gaussian mixture compressor and the domain feature resampler, SOYO efficiently stores and balances prior domain data without increasing memory usage or compromising data privacy. Our framework is universally compatible with various PIDIL tasks and supports multiple parameter-efficient fine-tuning methods. Extensive experiments on six benchmarks demonstrate that SOYO consistently outperforms existing baselines.

\section{Acknowledgments}
This work was funded by the National Natural Science Foundation of China under Grant No.U21B2048 and No.62302382, Shenzhen Key Technical Projects under Grant CJGJZD2022051714160501, China Postdoctoral Science Foundation No.2024M752584, and Natural Science Foundation of Shaanxi Province No.2024JC-YBQN-0637.

{
    \small
    \bibliographystyle{ieeenat_fullname}
    \bibliography{main}
}

\end{document}